\documentclass{article}

\usepackage{amsmath}
\usepackage{graphicx}
\usepackage{algorithm}
\usepackage{algorithmic}
\usepackage{float}

\providecommand{\norm}[1]{\lVert#1\rVert}

\begin{document}

\title{CloudSVM : Training an SVM Classifier in Cloud Computing Systems}
\author{F. Ozgur CATAK\\
National Research Institute Of Electronics and Cryptology (UEKAE) ,TUBITAK \\
\and
        M. Erdal BALABAN\\
        Industrial Engineering, Isik University
}

\date{}
\maketitle


\begin{abstract}
In conventional method, distributed support vector machines (SVM) algorithms are trained over pre-configured intranet/internet environments to find out an optimal classifier. These methods are very complicated and costly for large datasets. Hence, we propose a method that is referred as the Cloud SVM training mechanism (CloudSVM) in a cloud computing environment with MapReduce technique for distributed machine learning applications. Accordingly, (i) SVM algorithm is trained in distributed cloud storage servers 
that work concurrently; (ii) merge all support vectors in every trained cloud node; and (iii) iterate these two steps until the SVM converges to the optimal classifier function.
Large scale data sets are not possible to train using SVM algorithm on a single computer.
The results of this study are important for training of large scale data sets for machine learning applications.
We provided that iterative training of splitted data set in cloud computing environment using SVM will converge to a global optimal classifier in finite iteration size.
\end{abstract}

\section{Introduction}
Machine learning applications generally require large amounts of computation time and storage space. Learning algorithms have to be scaled up to handle extremely large data sets. When the training set is large, not all the examples can be loaded into memory in training phase of the machine learning algorithm at one step. It is required to distribute computation and memory requirements among several connected computers. \\

In machine learning field, support vector machines(SVM) offers most robust and accurate classification method due to their generalized properties. With its solid theoretical foundation and also proven effectiveness, SVM has contributed to researchers' success in many fields. 
But, SVM's suffer from a widely recognized scalability problem in both memory requirement and computational time\cite{Chang_psvm}.  SVM algorithm's computation and memory requirements increase rapidly with the number of instances in data set, many data sets are not suitable for classification\cite{Graf}. The SVM algorithm is formulated as quadratic optimization problem. Quadratic optimization problem has $O(m^3)$ time and $O(m^2)$ space complexity, where $m$ is the training set size\cite{tsang}. The computation time of SVM training is quadratic in the number of training instances. \\

The first approach to overcome large scale data set training is to reduce feature vector size. Feature selection and feature transformation methods are basic approaches for reducing vector size \cite{weston}. Feature selection algorithms choose a subset of the features from the original feature set and  feature transformation algorithms creates new data from the original feature space to a new space with reduced dimensionality. In literature, there are several methods; Singular Value Decomposition (SVD)\cite{golub}, Principal Component Analysis (PCA) \cite{jolliffe}, Independent Component Analysis (ICA)\cite{comon}, Correlation Based Feature Selection (CFS)\cite{hall}, Sampling based data set selection. All of these methods have a big problem for generalization of final machine learning model.\\

Second approach for large scale data set training is chunking \cite{Vapnik1995}.
Collobert et al. \cite{collobert} propose a parallel SVM training algorithm that each subset of whole dataset is trained with SVM and then the classifiers are combined into a final single classifier.
Lu et al.\cite{SvmStronglyConnected} proposed distributed support vector machine (DSVM) algorithm that finds support vectors (SVs) on strongly connected networks. Each site within a strongly connected network classifies subsets of training data locally via SVM and passes the calculated SVs to its descendant sites and receives SVs from its ancestor sites and recalculates the SVs and passes them to its descendant sites and so on.
Ruping et al.\cite{Ruping} proposed incremental learning with Support Vector Machine. One needs to make an error on the old Support Vectors(which represent the old learning set) more costly than an error on a new example.
Syed et al. \cite{Syed} proposed the distributed support vector machine (DSVM) algorithm that finds SVs locally and processes them altogether in a central processing center.
Caragea et al. \cite{Caragea} in 2005 improved this algorithm by allowing the data processing center to send support vectors back to the distributed data source and iteratively achieve the global optimum.
Graf et al. \cite{Graf} had an algorithm that implemented distributed processors into cascade top-down network topology, namely Cascade SVM. The bottom node of the network is the central processing center. The distributed SVM methods in these works converge and increase test accuracy.
All of these works have similar problems. They require a pre-defined network topology and computer size in their network. The performance of training depends on the special network configuration. Main idea of current distributed SVM methods is first data chunking then parallel implementation of SVM training. Global synchronization overheads are not considered in these approaches. \\

In this paper, we propose a Cloud Computing based SVM method with MapReduce \cite{MapReduceOrg} technique for distributed training phase of algorithm. By splitting training set over a cloud computing system's data nodes, each subset is optimized iteratively to find out a single global classifier. The basic idea behind this approach is to collect SVs from every optimized subset of training set at each cloud node, then merge them to save as global support vectors. Computers in cloud computing system exchange only minimum number of training set samples.  
Our algorithm CloudSVM is analysed with various public datasets. CloudSVM is built on the LibSVM and implemented using the Hadoop implementation of MapReduce.\\

This paper is organized as follows. In section 2, we will provide an overview to SVM formulations. In Section 3, presents the Map Reduce pattern in detail. Section 4 explains system model with our implementation of the Map Reduce pattern for the SVM training. In section 5, convergence of CloudSVM is explained. In section 6, simulation results with various UCI datasets are shown. Thereafter, we will give concluding remarks in Section 7.
\section{Support Vector Machine}
\begin{figure*}
	\centering
		\includegraphics[scale=0.4]{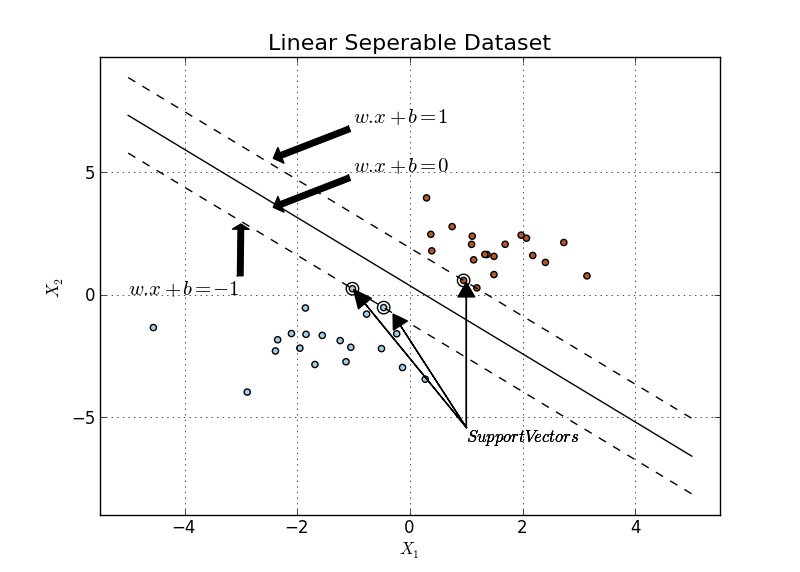}
	\caption{Binary classification of an SVM with Maximum-margin hyperplane trained with samples from two classes. Samples on the margin are called the support vectors.}
	\label{fig:linear_svm}
\end{figure*}

Support vector machine is a supervised learning method in statistics and computer science, to analyse data and recognize patterns, used for classification and regression analysis. The standard SVM takes a set of input data and predicts, for each given input, which of two possible classes forms the input, making the SVM a non-probabilistic binary linear classifier.
Note that if the training data[singular/plural] are linearly separable as shown in figure \ref{fig:linear_svm}, we can select the two hyperplanes of the margin in a way that there are no points between them and then try to maximize their distance. By using geometry, we find the distance between these two hyperplanes is $2 / \mathbf{\norm{w}}$. 
Given some training data $\mathcal{D}$, a set of n points of the form
\begin{equation} 
 \mathcal{D}=\{(\mathbf{x_i},y_i ) \; | \; \mathbf{x_i} \; \in \; R^m, \; y_i \; \in \; \{-1,1\} \;\}_{i=1}^n
\end{equation}
where $\mathbf{x_i}$ is an $m$-dimensional real vector, $y_i$ is either -1 or 1 denoting the class to which point $\mathbf{x_i}$ belongs.
SVMs aim to search a hyperplane in the
Reproducing Kernel Hilbert Space (RKHS) that maximizes the margin between the two classes of data in $\mathcal{D}$ with the smallest training error \cite{Vapnik1995}. This problem can be formulated as the following quadratic optimization
problem:
\begin{equation}\label{Eqn:QuadOpt}
\begin{split}
minimize : &  P(\mathbf{w},b,\xi) = \frac{1}{2} \norm{\mathbf{w}}^2 + C \sum_{i=1}^{m}{\xi_i}\\
subject to : & y_i(\left\langle \mathbf{w},\phi(\mathbf{x}_i) \right\rangle + b) \geq 1- \xi_i \\
      & \xi_i \geq 0
\end{split}
\end{equation}
for $i = 1,... , m$, where $\xi_i$ are slack variables and $C$ is a constant denoting the cost of each slack. $C$ is a trade-off parameter which controls the maximization of the margin and minimizing the training error. The decision function of SVMs is $f(\mathbf{x}) = \mathbf{w}^T \phi(\mathbf{x}) + b$ where the $\mathbf{w}$ and $b$ are obtained by solving the optimization problem $P$ in (\ref{Eqn:QuadOpt}).
By using Lagrange multipliers , the optimization problem $P$ in (\ref{Eqn:QuadOpt}) can be expressed as
\begin{equation}\label{Eqn:QuadOptLagrange}
\begin{split}
min : &  F(\mathbf{\alpha}) = \frac{1}{2} \mathbf{\alpha}^T \mathbf{Q} \mathbf{\alpha}^T - \mathbf{\alpha}^T\mathbf{1} \\
subject to : & \mathbf{0} \leq \mathbf{\alpha} \leq \mathbf{C}  \\
      & \mathbf{y}^T\alpha=0
\end{split}
\end{equation}

where $\left[ Q \right]_{ij} = y_i y_j \mathbf{\phi}^T(x_i)\mathbf{\phi}(x_j)$ is the Lagrangian multiplier variable. It is not need to know $\phi$, but it is necessary to know is how to compute the modified inner product which will be called as kernel function represented as
$K(\mathbf{x}_i,\mathbf{x}_j) = \phi^T(\mathbf{x}_i) \phi(\mathbf{x}_j)$.
Thus, $\left[ Q \right]_{ij} = y_iy_jK(x_i, x_j)$. 
Choosing a positive definite kernel K, by Mercers theorem, then optimization problem $P$ is a convex quadratic programming (QP) problem with linear constraints and can be solved in polynomial time.
\section{MapReduce}
MapReduce is a programming model derived from the map and reduce function combination from functional programming. MapReduce model widely used to run parallel applications for large scale data sets processing. Users specify a map function that processes a key/value pair to generate a set of intermediate key/value pairs, and a reduce function that merges all intermediate values associated with the same intermediate key\cite{MapReduceOrg}. MapReduce is divided into two major phases called map and reduce, separated by an internal shuffle phase of the intermediate results. The framework automatically executes those functions in parallel over any number of processors\cite{CloudBurst}. Simply, a MapReduce job executes three basic operations on a data set distributed across many shared-nothing cluster nodes. First task is Map function that processes in parallel manner by each node without transferring any data with other notes. In next operation, processed data by Map function is repartitioned across all nodes of the cluster. Lastly, Reduce task is executed in parallel manner by each node with partitioned data.
\begin{figure}[H]
	\centering
		\includegraphics[scale=0.5]{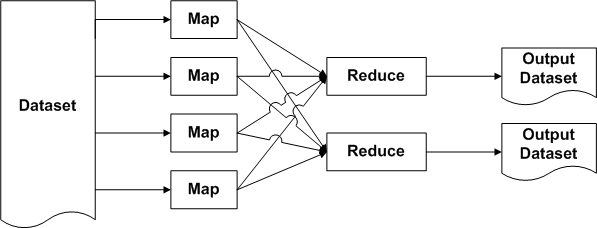}
	\caption{Overview of MapReduce System}
	\label{fig:MapReduceGeneral}
\end{figure}
A file in the distributed file system (DFS) is split into multiple chunks and each chunk is stored on different data-nodes.
A map function takes a key/value pair as input from input chunks and produces a list of key/value pairs as output. The type of output key and value can be different from input key and value: 
$$map(key_1,value_1) \Rightarrow list(key_2,value_2)$$
A reduce function takes a key and associated value list as input and generates a list of new values as output:
$$reduce(key_2,list(value_2)) \Rightarrow list(value_3) $$
Each Reduce call typically produces either one value $v_3$ or an empty return, though one call is allowed to return more than one value. The returns of all calls are collected as the desired result list.
Main advantage of MapReduce system is that it allows distributed processing of submitted job on the subset of a whole dataset in the network.
\section{System Model}
\begin{figure}[H]
	\centering
		\includegraphics[scale=0.5]{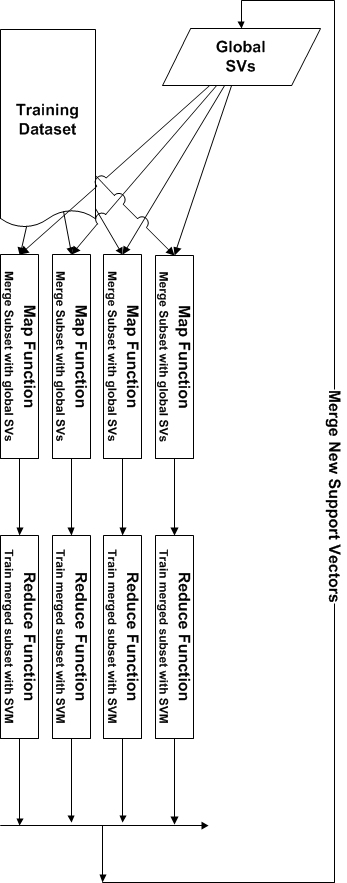}
	\caption{Schematic of Cloud SVM architecture.}
	\label{fig:SchematicCloudSVM}
\end{figure}
CloudSVM is a MapReduce based SVM training algorithm that runs in parallel on multiple commodity computers with Hadoop.
As shown in figure \ref{fig:SchematicCloudSVM}, the training set of the algorithm is split into subsets and each one is evaluated individually to get $\mathbf{\alpha}$ values (i.e. support vectors). In Map stage of MapReduce job, the subset of training set is combined with global support vectors. In Reduce step, the merged subset of training data is evaluated. The resulting new support vectors are combined with the global support vectors in Reduce step.
The CloudSVM with MapReduce algorithm can be explained as follows. First, each computer within a cloud computing system reads the global support vectors, then merges global SVs with subsets of local training data and classifies via SVM. Finally, all the computed SVs in cloud computers are merged. Thus, algorithm saves global SVs with new ones. The algorithm of CloudSVM consists of the following steps.
\begin{enumerate}
  \item As initialization the global support vector set as $t=0, V^t = \emptyset$
	\item t = t + 1;
	\item For any computer in $l,l=1,...,L$ reads global SVs and merge them with subset of training data.
	\item Train SVM algorithm with merged new data set
	\item Find out support vectors
	\item After all computers in cloud system complete their training phase, merge all calculated SVs and save the result to the global SVs
	\item If $h^t = h^{t-1}$ stop, otherwise go to step 2
\end{enumerate}

Pseudo code of CloudSVM Algorithm's Map and Reduce function are given in \textit{Algorithm \ref{Alg:CloudSVMMap}} and \textit{Algorithm \ref{Alg:CloudSVMReduce}}
\begin{algorithm}
\caption{Map Function of CloudSVM Algorithm}
\label{Alg:CloudSVMMap}
\begin{algorithmic}
\STATE $SV_{Global} = \emptyset$ \COMMENT{Empty global support vector set}
\WHILE{$h^t \neq h^{t-1}$}
\FOR{$l \in L$ \COMMENT{For each subset loop}} 
\STATE $\mathcal{D}_l^t \leftarrow \mathcal{D}_l^t \cup SV_{Global}^t$
\ENDFOR
\ENDWHILE
\end{algorithmic}
\end{algorithm}

\begin{algorithm}
\caption{Reduce Function of CloudSVM Algorithm}
\label{Alg:CloudSVMReduce}
\begin{algorithmic}
\WHILE{$h^t \neq h^{t-1}$}
\FOR{$l \in L$}
\STATE $SV_l,h^t \leftarrow svm(\mathcal{D}_l)$ \COMMENT{Train merged Dataset to obtain Support Vectors and Hypothesis }
\ENDFOR
\FOR{$l \in L$}
\STATE $SV_{Global} \leftarrow SV_{Global} \cup SV_l$
\ENDFOR
\ENDWHILE
\end{algorithmic}
\end{algorithm}

For training SVM classifier functions, we used LibSVM with various kernels. Appropriate parameters $C$ and $\gamma$ values were found by cross validation test. All system is implemented with Hadoop and streaming Python package mrjob library.
\section{Convergence of CloudSVM}
Let $\mathcal{S}$ denotes a subset of training set $\mathcal{D}$, $F(\mathcal{S})$ is the optimal objective function over data set $\mathcal{S}$, $h^*$ is the global optimal hypothesis for which has a minimal empirical risk $R_{emp}(h)$. Our algorithm starts with $\mathbf{SV}^0_{Global} = \mathbf{0}$, and generates a non-increasing sequence of positive set of vectors $\mathbf{SV}^t_{Global}$, where $\mathbf{SV}^t_{Global}$ is the vector of support vector at the $t$.th iteration. 
We used hinge loss for testing our models trained with CloudSVM algorithm. Hinge loss works well for its purposes in SVM as a classifier, since the more you violate the margin, the higher the penalty is\cite{HingeLoss}. The hinge loss function is the following:
$$l(f(x),y) = max\left\{0,1-y.f(x)\right\}$$
Empirical risk can be computed with an approximation:
$$R_{emp}(h) = \frac{1}{n}\sum_{i=1}^n{l(h(x_i),y_i)}$$
According to the empirical risk minimization principle the learning algorithm should choose a hypothesis $\hat{h}$ which minimizes the empirical risk:
$$\hat{h} = \arg \min_{h \in \mathcal{H}} R_{\mbox{emp}}(h).$$
A hypothesis is found in every cloud node. Let $\mathcal{X}$ be a subset of training data at cloud node $i$ where $\mathcal{X} \in R^{mxn}$, $\mathbf{SV}^t_{Global}$ is the vector of support vector at the $t.$th iteration, $h^{t,i}$ is hypothesis at node $i$ with iteration $t$, then the optimization problem in equation \ref{Eqn:QuadOptLagrange} becomes 
\begin{equation}\label{Eqn:ModQuadOptLagrange}
\begin{split}
maximize \, &h^{t,l} = -\frac{1}{2} 
\begin{bmatrix}
\mathbf{\alpha}_1 \\
\mathbf{\alpha}_2
\end{bmatrix}^T
\begin{bmatrix}
\mathbf{Q}_{11} & \mathbf{Q}_{12} \\
\mathbf{Q}_{21} & \mathbf{Q}_{22}
\end{bmatrix}
\begin{bmatrix}
\alpha_1 \\
\alpha_2
\end{bmatrix}
+
\begin{bmatrix}
1 \\
1
\end{bmatrix}^T
\begin{bmatrix}
\alpha_1 \\
\alpha_2 \\
\end{bmatrix}\\
subject to :\, & 0 \leq \alpha_i \leq C, \, \forall i \, and \, \sum_i^I{\alpha_iy_i}=0
\end{split}
\end{equation}

where $\mathbf{Q}_{12}$ and $\mathbf{Q}_{21}$ are kernel matrices with respect to 
$$ \mathbf{Q}_{12} = \left\{ K_{i,j}( x_{ij},SV_{Global(i,j)}^t ) \, | \, i=1,...,m , j=1,...,n \right\}.$$ 
$\mathbf{\alpha_1}$ and $\mathbf{\alpha_2}$ are the solutions estimated by node $i$ with dataset $\mathcal{X}$ and $\mathbf{SV}_{Global}$.
Because of the Mercer's theorem, our kernel matrix $Q$ is a symmetric positive-definite function on a square. Then our sub matrices $\mathbf{Q}_{12}$ and $\mathbf{Q}_{21}$ must be equal.\\
We can define $\mathbf{Q}_{11}$ and $\mathbf{Q}_{22}$ matrices such that
$$\mathbf{Q}_{11} = \left\{ K_{i,j}(x_{i,j},x_{i,j}) | x_{i,j} \in \mathcal{X}, i=1,...,m , j=1,...,n  \right\}$$
$$\mathbf{Q}_{22} = \left\{ K_{i,j}(SV_{Global},SV_{Global}) | i=1,...,m , j=1,...,n  \right\} $$ 
at iteration $t$. \\
Algorithm's stop point is reached when the hypothesis' empirical risk is same with previous iteration. That is:
\begin{equation}\label{Eqn:genAlgorithmStopPoint}
R_{emp}(h^t) = R_{emp}(h^{t-1})
\end{equation}
\textit{Lemma :} Accuracy of the decision function of CloudSVM classifier at iteration $t$ is always greater or equal to the maximum accuracy of the decision function of SVM classifier at iteration $t-1$. That is 
\begin{equation}\label{Eqn:AlgorithmStopPoint}
R_{emp}(h^t) \leq \arg \min_{h \in \mathcal{H}^{t-1}} R_{\mbox{emp}}(h) 
\end{equation}

\textit{Proof :}Without loss of generality, Iterated CloudSVM monotonically converges to optimum classifier.
$$\mathbf{SV}_{Global}^t = \mathbf{SV}_{Global}^{t-1} \cup \left\{ \mathbf{SV}_i^{t-1} \, | \, i = 1,...n \right\}$$ where $n$ is the data set split size(or cloud node size). Then, training set for svm algorithm at node $i$ is

$$d = \mathcal{X} \cup \mathbf{SV}_{Global}^t$$

Adding more samples cannot decrease the optimal value. Accuracy of the sub problem in each node monotonically increases in each step.

\section{Simulation Results}
\begin{table}[h]
\caption{The datasets used in experiments } 
\centering 
\begin{tabular}{crr rr rr} 
\hline
Dataset Name& Train. Data & Dim.   \\ [0.5ex]   
\hline             
German      & 1000 & 24 \\ 
Heart       & 270  & 13 \\
Ionosphere  & 351  & 34 \\
Satellite   & 4435 & 36 \\[1ex] 
\hline                          
\end{tabular}
\label{tab:dsInfo}
\end{table}

We have selected several data sets from the UCI Machine Learning Repository, namely, German, Heart, Ionosphere, Hand Digit and Satellite. The data sets length and input dimensions are shown in Table \ref{tab:dsInfo}. We test our algorithm over a real-word data sets to demonstrate the convergence. Linear kernels were used with optimal parameters ($\gamma , C$). Parameters were estimated by cross-validation method.

\begin{table}[h]
\caption{Performance Results of CloudSVM algorithm with various UCI datasets $\gamma$} 
\centering 
\begin{tabular}{c|r|r|r|r|r|r} 
\hline\hline                        
Dataset Name & $\gamma$ & $C$ &No. Of Iteration & No. of SVs &Accuracy&Kernel Type\\ [0.5ex]   
\hline             
German      & $10^0$ & 1 & 5 & 606 & 0.7728 & Linear\\ 
Heart       & $10^0$ & 1 & 3 & 137 & 0.8259 & Linear\\ 
Ionosphere  & $10^8$ & 1 & 3 & 160 & 0.8423 & Linear\\ 
Satellite   & $10^0$ & 1 & 2 & 1384 & 0.9064 & Linear\\[1ex] 
\hline                          
\end{tabular}
\label{tab:hresult}
\end{table}
We used 10-fold cross-validation, dividing the set of samples at random into 10 approximately equal-size parts. The 10 parts were roughly balanced, ensuring that the classes were distributed uniformly to each of the 10 parts. Ten-fold cross-validation works as follows: we fit the model on 90\% of the samples and then predict the class labels of the remaining 10\% (the test samples). This procedure is repeated 10 times, with each part playing the role of the test samples and the errors on all 10 parts added together to compute the overall error.\\
To analyse the CloudSVM, we randomly distributed all the training data to a cloud computing system with 10 computers with pseudo distributed Hadoop.\\
Data set prediction accuracy with iterations and total number of SVs are shown in Table \ref{tab:PredAccuracy}. 
When iteration size become 3 - 5, test accuracy values of all data sets reach to the highest values. If the iteration size is increased, the value of test accuracy falls into a steady state. The value of test accuracy is not changed for large enough number of iteration size.\\

\begin{table}[ht]
\caption{Data set prediction accuracy with iterations}
\centering
\begin{tabular}{cc}
\includegraphics[scale=0.5]{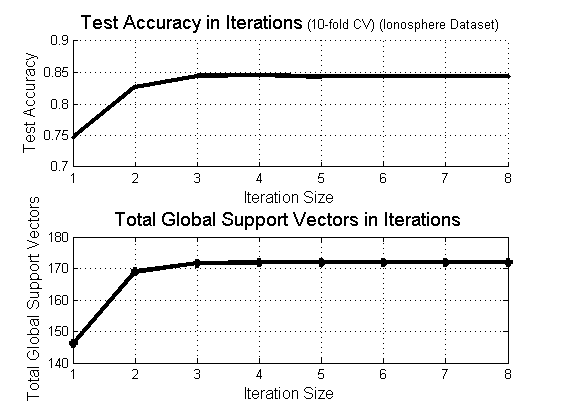} & \includegraphics[scale=0.5]{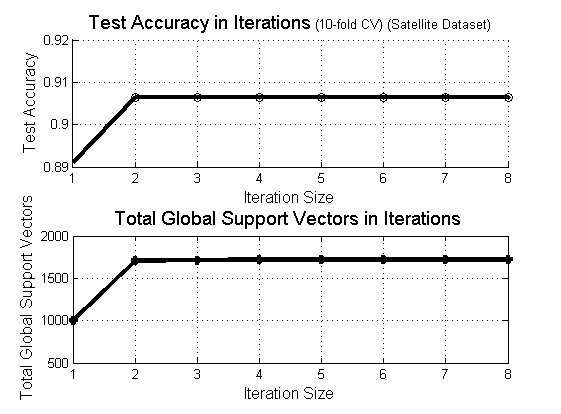}\\
\newline
\includegraphics[scale=0.5]{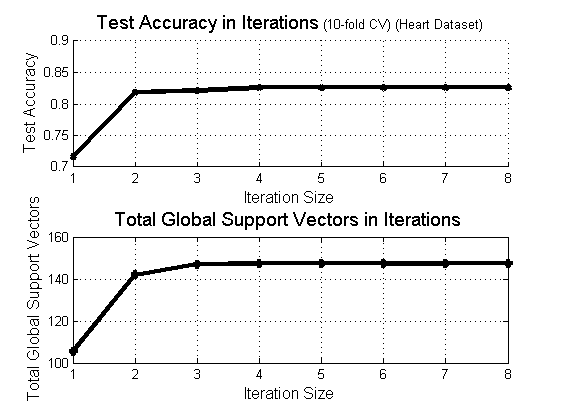}& \includegraphics[scale=0.5]{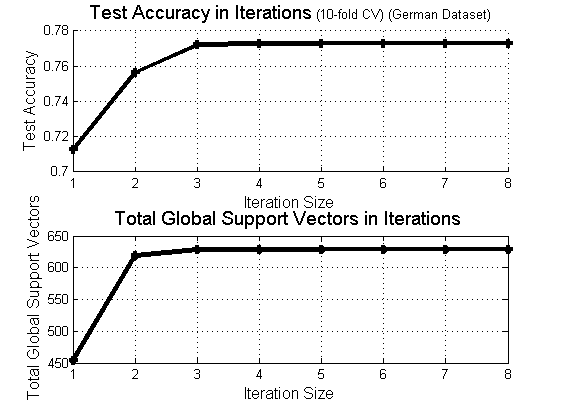}
\end{tabular}
\label{tab:PredAccuracy}
\end{table}

%
%
%
When the iteration size is increased, the number of global support vectors are passed the steady-state condition. As a result, the CloudSVM algorithm is useful for large size training data. 

\section{Conclusion and Further Research}
We have proposed distributed support vector machine implementation in cloud computing systems with MapReduce technique that improves scalability and parallelism of split data set training. The performance and generalization property of our algorithm are evaluated in Hadoop. Our algorithm is able to work on cloud computing systems without knowing how many computers connected to run parallel. The algorithm is designed to deal with large scale data set training problems. It is empirically shown that the generalization performance and the risk minimization of our algorithm are better than the previous results.\\

\end{document}